\documentclass{article}

\PassOptionsToPackage{numbers, compress}{natbib}

\usepackage[final]{neurips_2024}

\usepackage{amsmath} 
\usepackage[utf8]{inputenc} 
\usepackage[T1]{fontenc}    
\usepackage{url}            
\usepackage{booktabs}       
\usepackage{amsfonts}       
\usepackage{nicefrac}       
\usepackage{microtype}      
\usepackage[numbers]{natbib}
\usepackage[dvipsnames,table,xcdraw]{xcolor}
\definecolor{cvprblue}{rgb}{0.21,0.49,0.74}
\usepackage{graphicx}
\usepackage{caption}
\usepackage{arydshln}
\usepackage{tikz}
\usepackage{wrapfig}
\usepackage{float}
\usepackage{multirow}
\usepackage{pifont}
\usepackage{makecell}
\usepackage{subfigure}
\usepackage[pagebackref,breaklinks,colorlinks,citecolor=cvprblue]{hyperref}

\usepackage{mdframed}
\definecolor{mygray}{gray}{0.97}
\colorlet{shadecolor}{mygray}
\newmdenv[%
  backgroundcolor=mygray, 
  linewidth=0pt
]{newshaded}

\newcommand{\tablestyle}[2]{\setlength{\tabcolsep}{#1}\renewcommand{\arraystretch}{#2}\centering\small}

\title{Skywork R1V: Pioneering Multimodal Reasoning with Chain-of-Thought}

\author{
    \textnormal{Yi Peng}\thanks{Equal contribution}, \quad   \textnormal{Peiyu Wang}\footnotemark[1], \quad  \textnormal{Xiaokun Wang},  \quad  \textnormal{Yichen Wei},  \quad  \textnormal{Jiangbo Pei},  \quad  \textnormal{Weijie Qiu},  \\
    \textnormal{Ai Jian},    \quad \textnormal{Yunzhuo Hao},   \quad \textnormal{Jiachun Pan},  \quad  \textnormal{Tianyidan Xie}, \quad  \textnormal{Li Ge},  \\
    \textnormal{Rongxian Zhuang},  \quad   \textnormal{Xuchen Song}\thanks{Corresponding author}, \quad   \textnormal{Yang Liu}\footnotemark[2],  \quad     \textnormal{Yahui Zhou}\\
    \quad\\
    \textnormal{Skywork AI, Kunlun Inc.}\\
 {\textnormal{peiyu.wang@kunlun-inc.com, xuchen.song@kunlun-inc.com}}
}

\begin{document}

\maketitle

\begin{abstract}

We introduce Skywork R1V, a multimodal reasoning model extending the an R1-series Large language models (LLM) to visual modalities via an efficient multimodal transfer method. Leveraging a lightweight visual projector, Skywork R1V facilitates seamless multimodal adaptation without necessitating retraining of either the foundational language model or the vision encoder. To strengthen visual-text alignment, we propose a hybrid optimization strategy that combines Iterative Supervised Fine-Tuning (SFT) with Group Relative Policy Optimization (GRPO), significantly enhancing cross-modal integration efficiency. Additionally, we introduce an adaptive-length Chain-of-Thought distillation approach for reasoning data generation. This approach dynamically optimizes reasoning chain lengths, thereby enhancing inference efficiency and preventing excessive reasoning overthinking. Empirical evaluations demonstrate that Skywork R1V, with only 38B parameters, delivers competitive performance, achieving a score of 69.0 on the MMMU benchmark and 67.5 on MathVista. Meanwhile, it maintains robust textual reasoning performance, evidenced by impressive scores of 72.0 on AIME and 94.0 on MATH500. The Skywork R1V model weights have been publicly released to promote openness and reproducibility\footnote{\href{https://huggingface.co/Skywork/Skywork-R1V-38B}{https://huggingface.co/Skywork/Skywork-R1V-38B}}.

\end{abstract}

\section{Introduction}
In recent years, significant advances have been made in artificial intelligence, especially in natural language processing. Large language models (LLMs), represented by OpenAI GPT-4o \cite{openai2023gpt4}, Claude 3.5 \cite{Claude2024} and Deepseek-R1 \cite{deepseekai2025deepseekr1incentivizingreasoningcapability}, have achieved groundbreaking progress in complex reasoning tasks, reaching human-expert levels in logical reasoning and mathematical problem-solving within textual contexts. These models demonstrate proficiency in accurately interpreting complex problems, performing detailed step-by-step analyses, and ultimately arriving at correct solutions in intricate mathematical and logical reasoning tasks.

However, extending these advancements into multimodal contexts presents substantial challenges. Although vision-language models (VLMs) \cite{wang2023cogvlm,wang2023visionllm} excel at descriptive tasks—such as generating coherent and contextually relevant textual descriptions for images, their performance in deeply logical multimodal tasks (e.g., geometric proofs and scientific problem-solving) remains inferior to that of single-modal systems \cite{yang2023gpt4tools}. For example, For instance, geometric reasoning tasks demand that models accurately interpret intricate geometric relationships from visual inputs to carry out logical deductions effectively. Existing VLMs often struggle to accurately understand complex geometric relationships in images and conduct effective reasoning and proof.

Integrating reasoning-capable language models into VLMs to augment their reasoning capabilities presents a promising solution. Nevertheless, practical implementation of this integration faces significant obstacles. Specifically, the alignment between visual backbones and LLMs necessitates extensive datasets, while the unique nature of reasoning tasks demands specialized, reasoning-formatted training data. However, current VLM datasets predominantly consist of non-reasoning content, with only a limited subset containing traditional VLM chain-of-thought (CoT) \cite{wei2022chain} examples, which often lack the complexity needed for advanced reasoning tasks. Consequently, training VLMs on such datasets may inadvertently weaken rather than strengthen their multimodal reasoning capabilities.

To address these issues, we introduce Skywork R1V, a novel multimodal reasoning model. It transfers the reasoning capabilities of the R1 text model series to the visual domain via cross-modal transfer technology, achieving visual reasoning performance comparable to closed-source large models like Gemini2.0 \cite{geminiteam2024geminifamilyhighlycapable} and K1.5 \cite{team2025kimi}. owing to three core technical innovations:

\begin{enumerate}
\item \textbf{Efficient Multimodal Transfer of Reasoning-Capable LLMs}: Utilizing a lightweight multilayer perceptron (MLP) as a visual projector, Skywork R1V seamlessly transfers the reasoning prowess of R1-series text models into multimodal scenarios without retraining either the base language model or the visual encoder.

\item \textbf{Hybrid Optimization Framework}: This framework strategically integrates Iterative Supervised Fine-Tuning (SFT) with Group Relative Policy Optimization (GRPO) reinforcement learning, progressively aligning visual-textual representations for efficient cross-modal reasoning. 

\item \textbf{Adaptive-Length Chain-of-Thought Distillation}: By dynamically adjusting the length of reasoning chains, this technology mitigates excessive computational deliberation overthinking, significantly enhancing reasoning efficiency and inference effectiveness.
\end{enumerate}

All components and weights of Skywork R1V have been fully open-sourced, aiming to foster broader research and innovation within the multimodal reasoning community.

\section{Methodology}
In this section, we elaborate on the technical details of Skywork R1V. Section \ref{sec:multimodal_transfer} describes our approach for transferring a reasoning-capable LLM into the MLP-based VLM framework. Section \ref{sec:train} outlines our training framework, and Section \ref{sec:cot_gene} provides detailed insights into the training data generation process.

\subsection{Efficient Multimodal Transfer of Reasoning-Capable LLMs}

\label{sec:multimodal_transfer}

We propose an Efficient Multimodal Transfer method, efficiently aligning a reasoning-capable language model with a vision backbone through an MLP structure, substantially reducing the requirement for extensive multimodal reasoning data.

The insight behind our approach lies in decoupling the alignment of visual-language representations from the preservation of reasoning capabilities. Directly connecting the reasoning-capable language model (\(f_l\)) to a vision backbone (\(f_v\)) would necessitate extensive multimodal reasoning data in the R1-style format to simultaneously achieve both objectives. However, acquiring such data is prohibitively expensive and impractical for most applications.

Instead, we adopt a staged strategy. First, we train an MLP adapter to align \(f_v\) with a substitutive language model (\(f_l^{s}\)) sharing the same architecture as \(f_l\) but without reasoning capabilities. This initial step allows the MLP to learn a generalized mapping from the visual space of \(f_v\) to the language space of \(f_l^{s}\), utilizing existing multimodal datasets. Subsequently, leveraging the latent similarity between \(f_l^{s}\) and \(f_l\), transferring this pretrained MLP to align \(f_v\) with the original reasoning-capable model \(f_l\) becomes significantly more efficient, requiring substantially fewer data. Our method is detailed further in the following three steps.

\paragraph{MLP Initialization} 
Given a vision encoder \( f_v \) (we employ the \texttt{Vision Transformer (ViT)} \cite{dosovitskiy2020vit}), a reasoning-capable language model \( f_l \) (\texttt{DeepSeek-R1-distill-Qwen2.5-32B} \cite{deepseekai2025deepseekr1incentivizingreasoningcapability}), and a substitutive language model \( f_l^s \) (\texttt{Qwen2.5-32B-Instruct} \cite{qwen2.5}), we first initialize the MLP adapter by aligning \( f_v \) with \( f_l^s \). Specifically, the MLP adapter \( \theta \) connects \( f_v \) and \( f_l^s \), forming a preliminary vision-language model \( M' = f_v \circ \theta \circ f_l^s \). Keeping both \( f_v \) and \( f_l^s \) frozen, we optimize the MLP parameters through the following SFT process: 1) initial fine-tuning on the full dataset (2M samples); 2) refinement on a curated subset consisting of 200K high-quality samples selected via GPT-4 evaluation; and 3) a final fine-tuning step on 40K high-quality Chain-of-Thought (CoT) samples. The learning rate is set to \(2\times10^{-4}\) for the initial fine-tuning stage, and reduced to \(4\times10^{-5}\) for the subsequent refinement stages (2 and 3). Additional hyperparameters include a context length of 16384 tokens, a weight decay of 0.05, a warmup ratio of 0.03, a batch size of 512, and a training epoch of 1 for each stage.

\paragraph{Model Re-Assembly} 
Using the pretrained MLP adapter \( \theta \) obtained from Step 1, we transfer it to bridge the vision encoder \( f_v \) and the original reasoning-capable language model \( f_l \), thus constructing the complete R1V model \( M = f_v \circ \theta \circ f_l \). The tokenizer for the combined model follows that of \( f_l \). \textbf{Notably}, despite the changes in both the language model parameters and the tokenizer, the assembled model surprisingly retains a significant portion of its original performance. For further details on this phenomenon, please refer to Section~\ref{sec:exp_model_reass}.

\paragraph{Modality Alignment} 
Finally, we perform modality alignment between the visual and textual representations within the model \( M \). During this phase, only the MLP adapter parameters \( \theta \) are fine-tuned, while both the vision encoder \( f_v \) and the reasoning-capable language model \( f_l \) remain fixed. This approach ensures the model retains robust reasoning abilities inherited from the R1-series LLM, while effectively aligning visual and language modalities. This alignment is carried out via the Hybrid Optimization Framework detailed in Section~\ref{sec:train}, utilizing training data in the form of reasoning chains generated through our Adaptive-Length Chain-of-Thought method as described in Section~\ref{sec:cot_gene}.

\begin{figure*}
    \centering
    \includegraphics[width=1\linewidth]{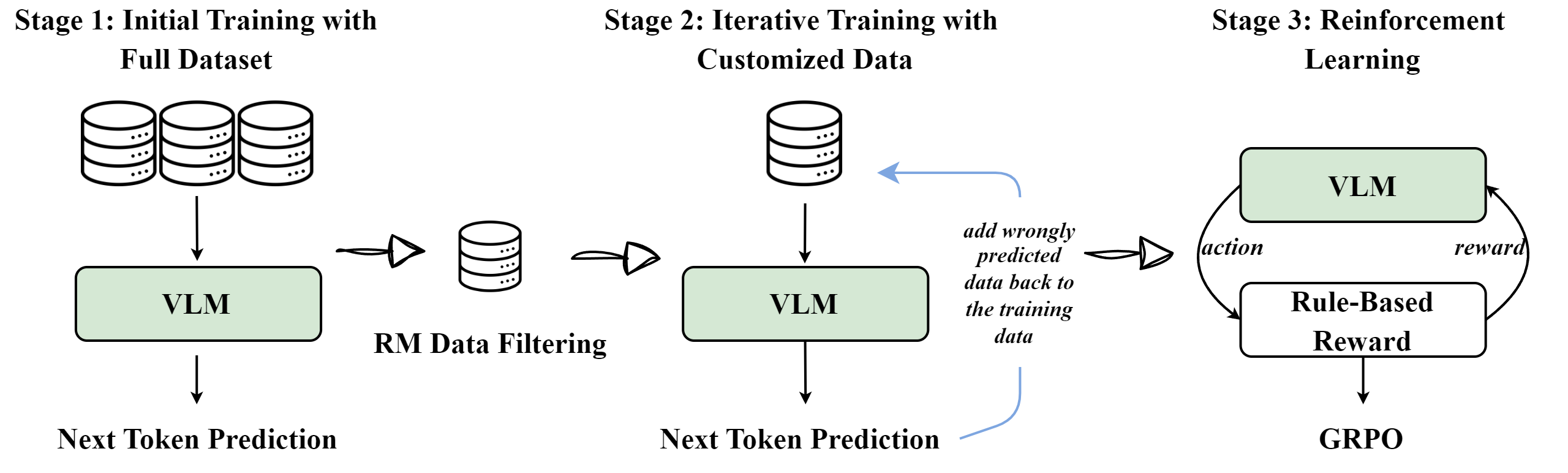}
    \caption{Hybrid Optimization Framework.} 
    \label{fig: train} 
\end{figure*} 

\subsection{Hybrid Optimization Framework}
\label{sec:train}

We propose a Hybrid Optimization Framework (Figure \ref{fig: train}) that strategically integrates iterative SFT and GRPO. Specifically, during the iterative SFT phase, we sequentially train a series of models \( M_0, M_1, \dots, M_T \). Each subsequent model \( M_{t+1} \) is trained using high-quality data identified by our reward model, along with challenging samples misclassified by the previous iteration. In the RL phase, we apply GRPO to further enhance model generalization

\paragraph{Stage 1: Initial Training with Full Dataset}
We commence by performing SFT of the base model using our full dataset $\mathcal{D}$. The training configuration follows the aforementioned \textbf{MLP Initialization}. The data generation process is introduced in Section~\ref{sec:cot_gene}. This stage produces the initialized model $M_0$.

\paragraph{Stage 2: Iterative Training with Customized Data}
During the iterative SFT phase, we sequentially train a series of models \( \{M_1, M_2, \dots, M_T\} \) (with \(T=4\)). Each model \( M_{t} \) is iteratively fine-tuned based on its predecessor \( M_{t-1} \), leveraging high-quality samples identified by our Reward Model (RM), as well as challenging cases incorrectly processed by \( M_{t-1} \). Specifically, the Reward Model assigns a quality score to each sample as follows:
\[
\text{RM}: \mathcal{X} \to \{0, 1, 2, 3, 4, 5\}.
\]

Utilizing these scores, we construct a refined dataset \(\mathcal{D}_{\text{rm}}\) by selecting high-quality samples from the original dataset \(\mathcal{D}\) according to a dynamically increasing threshold:
\[
\mathcal{D}_{\text{rm}} = \{(x,y) \in \mathcal{D} \mid \text{RM}(x) \geq \tau\},
\]
where the threshold \(\tau\) is progressively set to 2, 3, 4, and 5 for iterations \(t=1,2,3,4\), respectively.

Additionally, for each iteration \(t\), we construct an error-focused dataset \(\mathcal{E}_{t-1}\), explicitly targeting challenging samples that were misclassified in the previous iteration:
\[
\mathcal{E}_{t-1} = \{ (x,y) \in \mathcal{D} \mid \Phi(M_{t-1}(x)) \neq y \},
\]
where \(M_{t-1}(x)\) denotes the response from the model at iteration \(t-1\), \(y\) represents the ground-truth label, and \(\Phi\) is a function used to extract answers from the model's outputs.

The combined, customized training dataset for each iteration \(t\) is thus formulated as:
\[
\mathcal{D}_t = \mathcal{D}_{\text{rm}} \cup \mathcal{E}_{t-1}.
\]

Finally, we fine-tune the previous iteration's model \( M_{t-1} \) using the customized dataset \(\mathcal{D}_t\), thus enhancing the model's robustness and generalization capabilities, resulting in the improved model \( M_{t} \). For each iteration, we train the model for 1 epoch, employing a context length of 16,384 tokens, weight decay of 0.05, warmup ratio of 0.03, and  batch size of 512. The learning rate is set to \(1\times10^{-4}\) for the first iteration and subsequently reduced to \(2\times10^{-5}\) for the following iterations.

\paragraph{Stage 3:  Reinforcement Learning}
Following the approach proposed in DeepSeek R1, we utilize the GRPO with rule-based reward system (Accuracy reward $\&$ Format reward) to  further boost the generalizability of our model. The reward-model-filtered subset \(\mathcal{D}_{\text{rm}}\) (\(\tau = 5\)) is utilized as the training dataset.  The model is trained with the following hyperparameters: learning rate of $1\times10^{-6}$, temperature of 1.0, generation number of 8, and a maximum completion length of 8k tokens.  After RL training, we select the model that achieves the optimal balance between performance and reasoning rationality, designating it as the final model.

\subsection{Adaptive-Length Chain-of-Thought Distillation}
\label{sec:cot_gene}

\begin{figure*}
    \centering
    \includegraphics[width=1\linewidth]{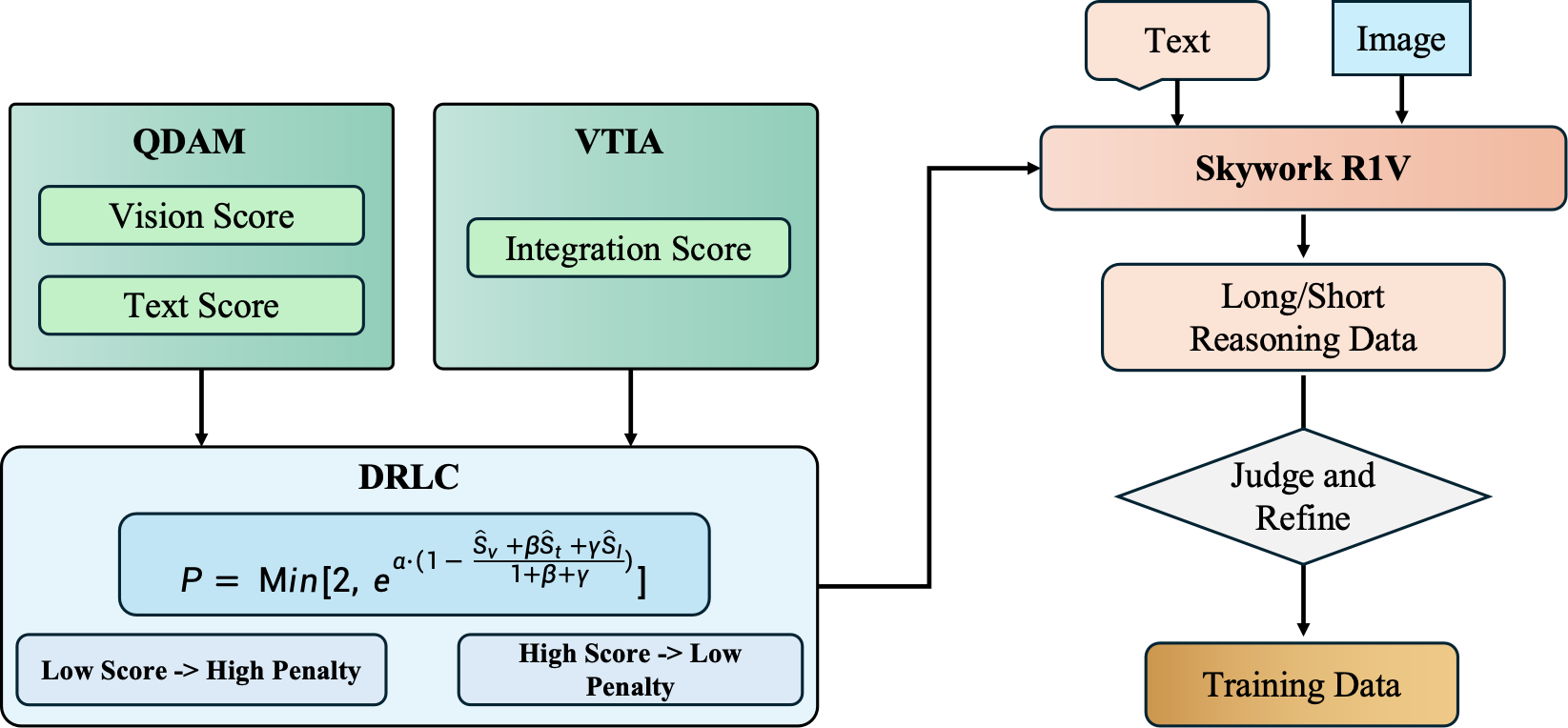}
    \caption{Adaptive-Length Chain-of-Thought Distillation.} 
    \label{fig: data} 
\end{figure*}

We propose an Adaptive-Length Chain-of-Thought Distillation (AL-CoTD) framework (Figure \ref{fig: data}), specifically designed to dynamically optimize the reasoning chain length when generating high-quality reasoning-oriented data. The generated data effectively mitigates the common issue of excessive reasoning or overthinking during inference.

\paragraph{Quality and Difficulty Assessment Module (QDAM)} 
The QDAM leverages GPT-4o to systematically evaluate image-text query pairs across two primary dimensions: the \textit{vision score} ($S_v$) and the \textit{text score} ($S_t$). Specifically, the \textit{vision score} assesses visual characteristics via two criteria—\textit{image clarity} and \textit{image necessity}. \textit{Image clarity} quantifies perceptual quality using blur detection and resolution analysis, whereas \textit{image necessity} evaluates the dependency of the text on visual context through context ablation tests and relevance classification. The \textit{text score} examines linguistic properties through three distinct aspects: \textit{question quality}, which assesses clarity using grammatical validation and semantic coherence checks; \textit{difficulty level}, which measures conceptual complexity based on domain-specific knowledge requirements; and \textit{reasoning demand}, which quantifies the complexity of inference steps via multi-hop reasoning analysis. Together, these measures offer a comprehensive framework for capturing both perceptual and cognitive complexities inherent in multimodal query understanding. All these properties are obtained by using GPT-4o, except for image clarity.

\paragraph{Vision-Text Integration Analyzer (VTIA)} 
VTIA quantifies the required depth of cross-modal integration by performing syntactic and semantic analyses, generating an \textit{integration score} ($S_I$) through pattern recognition within image-text queries using GPT-4o. Queries with high integration patterns, resulting in increased $S_I$, are typically found in tasks demanding scientific explanations or detailed reasoning. Such patterns include the presence of causal connectives (\textit{``why''}/\textit{``how''}) accompanied by presupposition triggers, multiple-object visual references necessitating spatial relationship comprehension, and co-occurrence of domain-specific terminologies. Conversely, queries exhibiting low integration patterns lead to reduced $S_I$. These are commonly seen in simpler tasks like object recognition, characterized by straightforward interrogatives (\textit{``what''}/\textit{``where''}) accompanied by definite articles, queries targeting direct object identification, and minimal dependence between the textual content and visual input. This pattern-driven analytical framework facilitates adaptive cross-modal fusion precisely tailored to the complexity of each query.

\textbf{Dynamic Reasoning Length Controller (DRLC)}  
The DRLC module operates on normalized scores $\hat{S}_v$, $\hat{S}_t$, and $\hat{S}_I$, derived from the original scores $S_v$, $S_t$, and $S_I$ via min-max scaling to the range $[0,1]$. The controller dynamically adjusts the reasoning chain length by modulating the \textit{repetition penalty} based on query complexity. Specifically, queries characterized by high visual-textual quality ($\hat{S}_v$, $\hat{S}_t$), substantial cognitive difficulty, and complex visual scenarios demanding deeper reasoning (reflected by higher values of $\hat{S}_v$, $\hat{S}_t$, and $\hat{S}_I$) receive a lower repetition penalty, allowing for longer reasoning chains. Conversely, queries of lower difficulty, simpler visual identification tasks, and minimal cross-modal integration requirements (indicated by lower $\hat{S}_v$, $\hat{S}_t$, and $\hat{S}_I$ scores) are assigned higher repetition penalties to prevent unnecessary reasoning. The repetition penalty ($P$) is calculated as:

\begin{equation}
    P = \min \left[ 2,\; e^{\alpha \cdot \left(1-\frac{\hat{S}_v+\beta \hat{S}_t+\gamma \hat{S}_I}{1+\beta+\gamma}\right)} \right],
\end{equation}

where $\alpha$, $\beta$, and $\gamma$ are hyperparameters controlling the relative influence of these components.

\paragraph{Multi-Stage Self-Distillation Pipeline} 
Building upon our DRLC module, we further propose a progressive self-distillation strategy. In this pipeline, the model initially generates reasoning-oriented data explicitly annotated with \texttt{<think>} tokens, where the repetition penalty \(P\), computed by the DRLC module, dynamically regulates the inference length. Subsequently, GPT-4o evaluates the correctness of the generated answers. If an answer is assessed as correct, the original reasoning chain (\texttt{<think>} annotations) is preserved; otherwise, GPT-4o revises the reasoning process to realign it with the correct answer. This  procedure is conducted prior to Stage 1 and repeated before each iteration of Stage 2 to refine the reasoning chain within our Hybrid Optimization Framework.

\section{Experiments}
We conducted a comprehensive evaluation of our model across multiple benchmarks designed to assess performance in different modalities. The benchmarks primarily fall into two categories:

\begin{table}[h]

    \centering
    \resizebox{.99\textwidth}{!}{
    \tablestyle{8pt}{1.1}
    \centering
    \small
    \begin{tabular}{@{}c l  c c  c  c c c@{}}
    \toprule
    
    & \multirow{3}{*}{\centering \textbf{Benchmark}} 
    & \multicolumn{1}{c}{\textbf{LLM}} 
    & \multicolumn{5}{c}{\textbf{VLM}} \\ 
    
    \cmidrule(rl){3-3} \cmidrule(rl){4-8}
    
    & &      \textbf{QwQ-32B} & 
    \textbf{QwenVL} &
    \textbf{InternVL-} & \textbf{VILA} & \textbf{InternVL2} & \textbf{Skywork} \\
    & & \textbf{-Preview} &   \textbf{2-72B} & \textbf{2.5-38B} & \textbf{1.5-40B} & \textbf{-40B} & \textbf{R1V (38B)}  \\
    \midrule

    \multirow{4}{*}{Reasoning}
    & MATH-500  & 90.6 & - & - & - & - & \textbf{94.0} \\
    & AIME 2024  & 50.0 & - & - & - & - & \textbf{72.0} \\
    & GPQA  {\tiny } & 54.5 & - & - & - & - & \textbf{61.6} \\
    \midrule

    \multirow{2}{*}{Vision} 
    & MathVista{\tiny(mini)} & - & 70.5 & \textbf{71.9} & 49.5 & 63.7 &67.5  \\
    & MMMU{\tiny(Val)} & - & 64.5 & 63.9 & 55.1 & 55.2 &\textbf{69.0}  \\

    \bottomrule
    \end{tabular}
}  
    \vspace{1mm}
     \caption{Evaluation results of state-of-the-art LLMs and VLMs.}
    \label{tab:table2}
\end{table}

\begin{itemize}
   
    \item \textbf{Reasoning Benchmarks}: 
    \begin{enumerate}
        \item \textbf{MATH-500} \citep{hendrycks2021measuring}: This dataset comprises 500 undergraduate-level mathematical problems spanning algebra, calculus, probability, and various other topics. It evaluates both computational proficiency and advanced mathematical reasoning, with higher scores reflecting superior problem-solving abilities.

        \item \textbf{AIME 2024}: This benchmark includes competition problems from the 2024 American Invitational Mathematics Examination (AIME), a prestigious and highly selective contest for elite high school students. It assesses advanced mathematical competencies, requiring deep conceptual understanding and rigorous logical reasoning skills.

        \item \textbf{GPQA} \cite{rein2024gpqa}: GPQA evaluates the general-purpose question-answering capabilities of language models. It comprises carefully designed questions spanning diverse domains, providing a robust measure of a model's ability to comprehend, analyze, and accurately respond to complex queries across multiple knowledge areas.
    \end{enumerate}

    \item \textbf{VLM Benchmarks}:
    \begin{enumerate}

        \item \textbf{MathVista} \citep{lu2023mathvista}: MathVista presents challenges integrating mathematical reasoning and visual understanding. It combines diverse tasks requiring precise visual interpretation and structured analytical reasoning, thus evaluating a model's capability to handle intricate multimodal problems.

        \item \textbf{MMMU} \citep{yue2024mmmu}: MMMU consists of approximately 11,500 questions sourced from college-level exams, quizzes, and textbooks, covering six academic disciplines: Art \& Design, Business, Science, Health \& Medicine, Humanities \& Social Science, and Tech \& Engineering. It assesses the model's proficiency in comprehending and responding effectively to complex multimodal inputs.
        
    \end{enumerate} 

\end{itemize}

\paragraph{Evaluation Setup}

In our evaluations, the maximum generation length is set to 64K tokens. For textual reasoning benchmarks, the test prompts strictly adhere to the implementation guidelines provided by DeepseekR1. For visual-language model (VLM) benchmarks, including MMMU, and MathVista, we utilize a unified test prompt. The reported performance metric is the Pass@1 score, averaged across 5 independent runs.

\begin{newshaded}
\small
\noindent \ding{227} {\bf Prompt for  Multi-Choice QA Problems}: 
Answer the multiple choice preceding question. The last line of your response should follow this format: 'Answer: $\backslash$boxed\{\$LETTER\}' (without quotes), where LETTER is one of the options. If you are uncertain or the problem is too complex, make a reasoned guess based on the information provided. Avoid repeating steps indefinitely—provide your best guess even if unsure. Think step by step logically, considering all relevant information before answering.

\noindent \ding{227} {\bf Prompt for Other Problems}: 
Answer the preceding question. The last line of your response should follow this format: 'Answer: $\backslash$boxed\{\$FINAL\_ANSWER\}' (without quotes), where FINAL\_ANSWER is is your conclusion. If you are uncertain or the problem is too complex, make a reasoned guess based on the information provided. Avoid repeating steps indefinitely—provide your best guess even if unsure. Think step by step logically, considering all relevant information before answering.
\end{newshaded}

\textbf{Baselines} We conduct comprehensive evaluations against several strong closed-source models, including Claude-3.5-Sonnet (20241022) \cite{Claude2024}, GPT-4o-0513 \cite{openai2024gpt4o}, OpenAI-o1-mini \cite{jaech2024openai}, and Kimi k1.5 \cite{team2025kimi}. Additionally, we compare our method with advanced open-source models, such as 
InternVL2-40B \cite{chen2024expanding}, InternVL2.5-38B \cite{chen2024expanding}, InternVL2.5-78B \cite{chen2024expanding}, VILA-1.5-40B \cite{lin2023vila}, QwQ-32B-Preview \cite{qwq-32b-preview} , Deepseek V3 \cite{deepseekai2024deepseekv3technicalreport}, Deepseek R1 \cite{deepseekai2025deepseekr1incentivizingreasoningcapability}, Qwen2-VL-72B-Instruct \cite{Qwen2VL} and Qwen2.5-VL-72B-Instruct \cite{bai2025qwen2}.

\begin{figure*}
    \centering
    \includegraphics[width=1\linewidth]{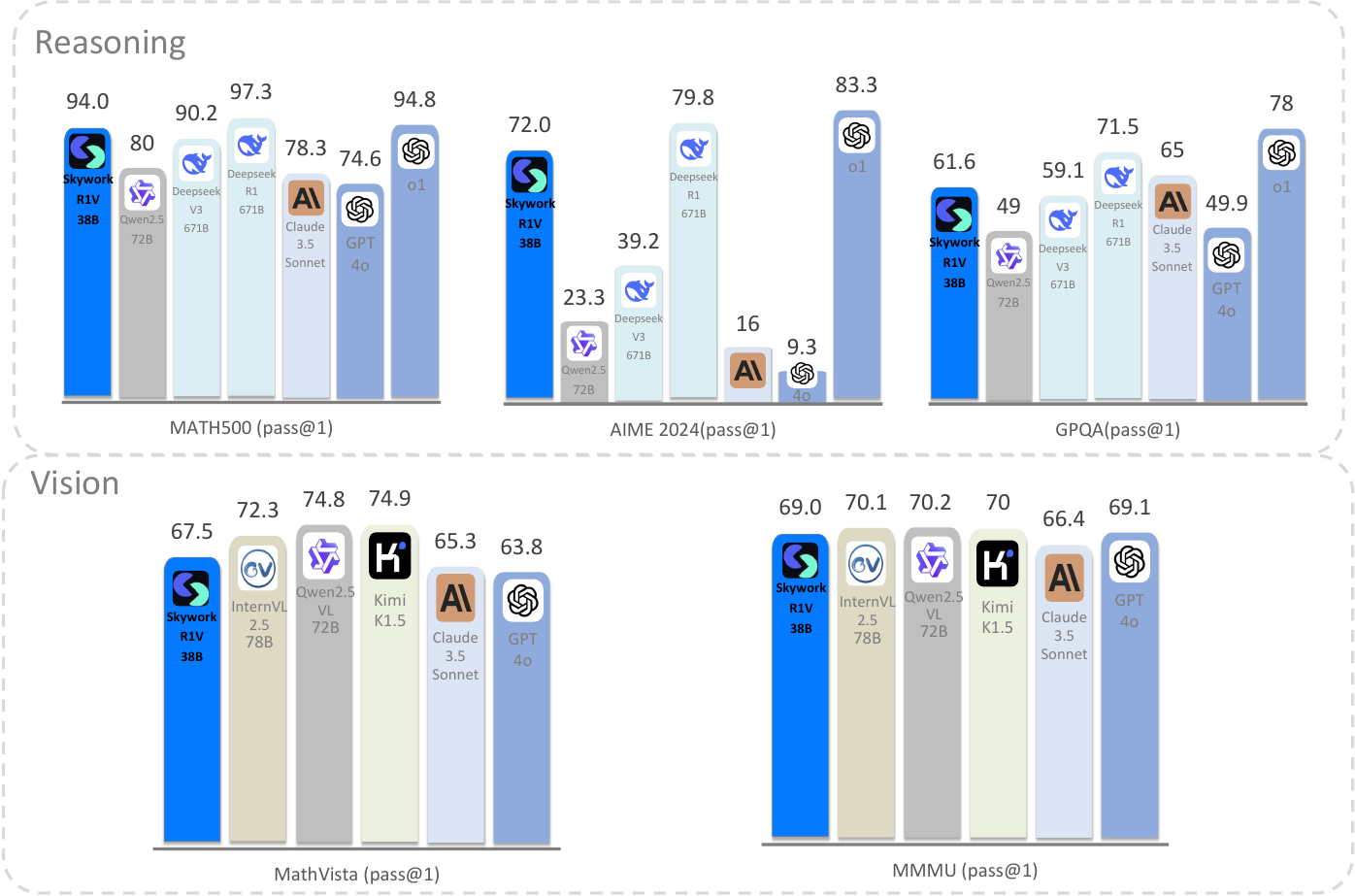}
    \caption{Comparison with Larger-Scale Open-Source and Closed-Source Models.} 
    \label{fig:pr2} 
\end{figure*}

\subsection{Main Results}
\paragraph{Comparison with Models of Similar Scale}
We comprehensively compare the performance of our Skywork R1V model with other state-of-the-art models of similar scale across various benchmarks. As shown in the updated evaluation results (Table~\ref{tab:table2}), Skywork R1V exhibits outstanding performance in both reasoning and visual tasks.

In text-based reasoning tasks, Skywork R1V achieves exceptional results, notably scoring 94.0 on the MATH-500 benchmark, surpassing similar-scale models such as QwQ-32B-Preview (90.6), and demonstrating significant advantages on the AIME 2024 benchmark with a remarkable score of 72.0.

In visual multimodal tasks, Skywork R1V also demonstrates robust performance. Specifically, on the MathVista (mini) benchmark, it achieves a score of 67.5, maintaining a strong performance relative to InternVL2-40B (63.7), though trailing InternVL2.5-38B (71.9). Moreover, on the MMMU (Val) benchmark, Skywork R1V attains a Pass@1 score of 69.0, significantly surpassing  VILA-40B (55.1), InternVL2-40B (55.2), InternVL2.5-38B (63.9), and QwenVL2-72B (64.5). These results collectively underscore the superior capability of Skywork R1V across a wide spectrum of tasks, affirming its effectiveness in both advanced reasoning and complex visual understanding scenarios.

\paragraph{Comparison with Larger-Scale Open-Source and Closed-Source Models}
We further compare our Skywork R1V model with larger-scale open-source and closed-source models (Figure \ref{fig:pr2}). Despite being only a 38B-scale model, Skywork R1V achieves competitive performance compared to significantly larger counterparts. Specifically, on the MathVista benchmark, our model achieves a score of 67.5, surpassing the performance of some closed-source models such as Claude 3.5 Sonnet (65.3). More notably, on the MMMU benchmark, Skywork R1V attains a score of 69.0, outperforming Claude 3.5 Sonnet (66.4) and matching GPT4-o (69.1). These results indicate that our approach effectively transfers advanced reasoning capabilities from textual modalities to vision, enabling our smaller-scale model to perform on par with larger-scale models.

\begin{figure*}
    \centering
    \includegraphics[width=1\linewidth]{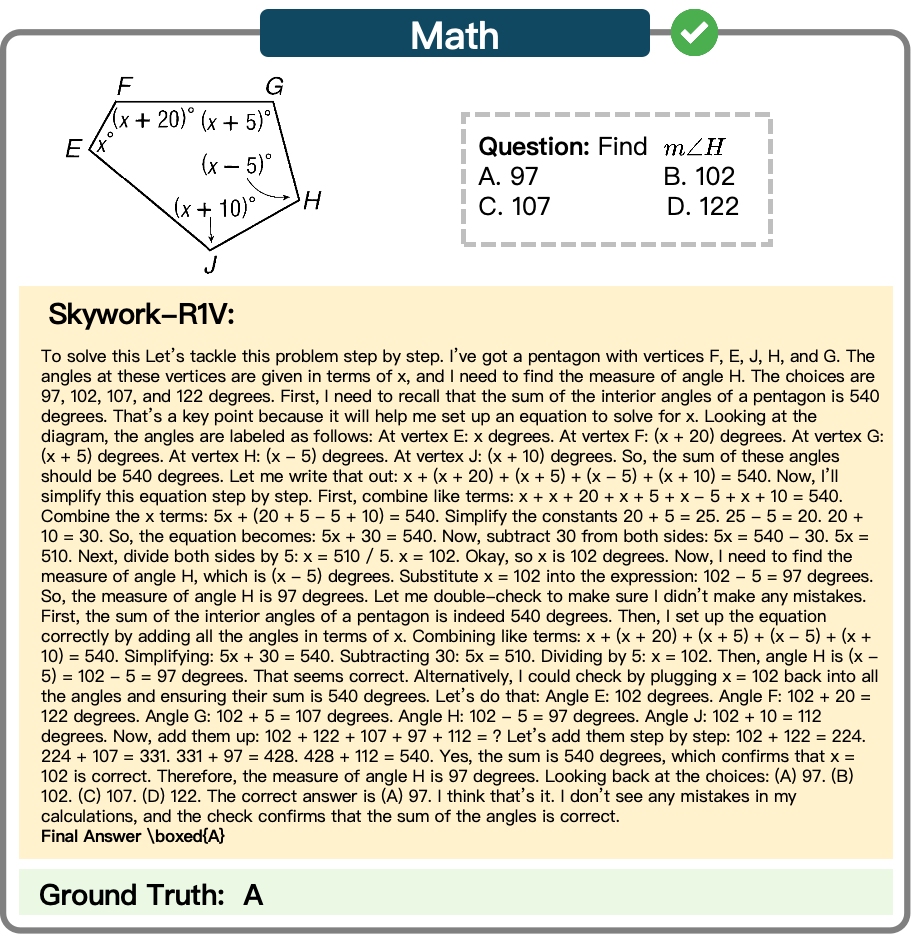}
    \caption{Reasoning Capability of Skywork R1V on Mathematical Problems.} 
    \label{fig:case1} 
\end{figure*}

\begin{figure*}
    \centering
    \includegraphics[width=1\linewidth]{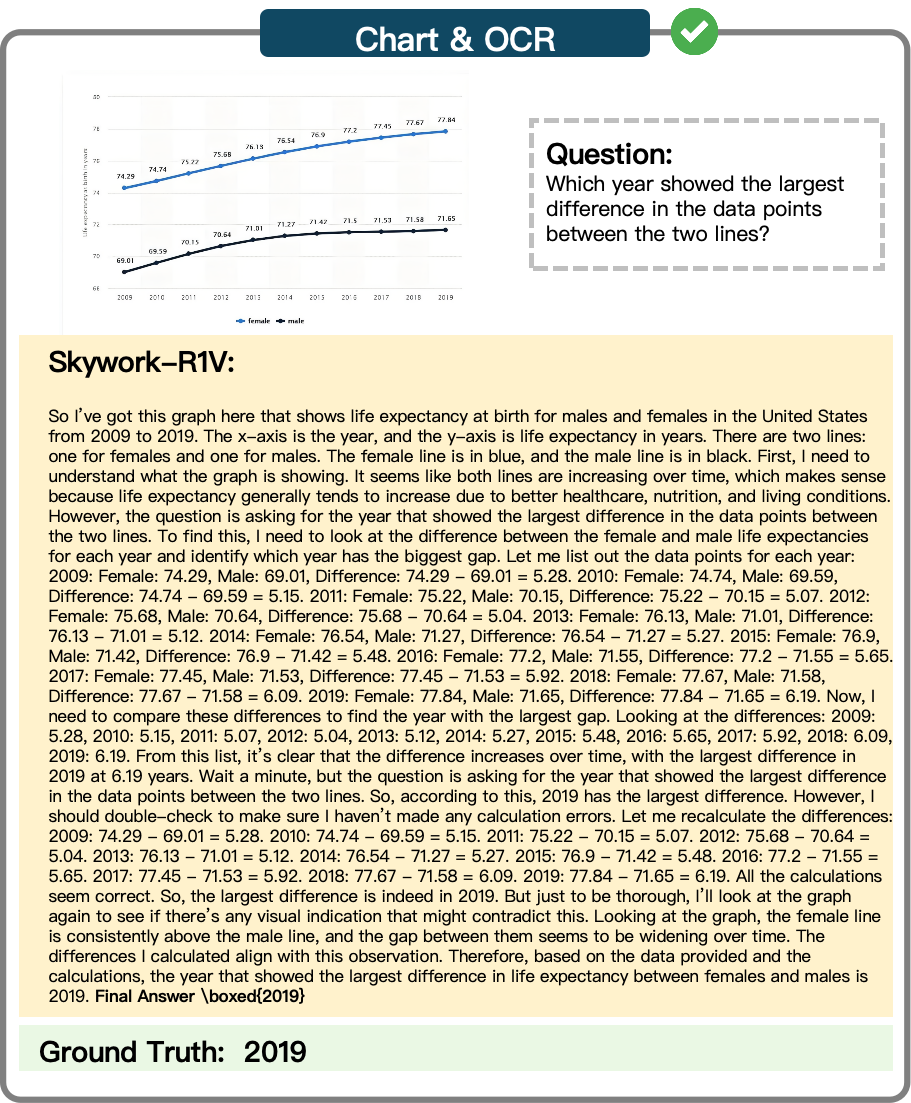}
    \caption{Reasoning Capability of Skywork R1V on Chart Problems.} 
    \label{fig:case2} 
\end{figure*}

\subsection{Analysis}  
\paragraph{Analysis of Reasoning Capability}  
As shown in Figure~\ref{fig:case1}, Skywork R1V addresses the pentagon angle problem with structured mathematical reasoning. It first applies the geometric principle that a pentagon’s interior angles sum to \(540^\circ\), constructs a linear equation from symbolic angle expressions, and solves for \(x = 102\) through algebraic simplification. The model then validates its solution by substituting \(x\) back into all angles to confirm the total equals \(540^\circ\), and calculates angle \(H = 97^\circ\) to align with contextual objectives. This dual-phase approach—systematic problem-solving coupled with self-verification—demonstrates rigorous integration of geometric and algebraic reasoning, critical for complex mathematical tasks.

Figure \ref{fig:case2} tests the model’s reasoning capabilities through a chart analysis task. The model begins by accurately interpreting the graph structure, identifying axes, gender-specific trends, and temporal patterns in U.S. life expectancy data (2009–2019). To pinpoint the year with the largest gender gap, it systematically calculates yearly differences between female and male values, validates results through recalculation to eliminate arithmetic errors, and cross-checks numerical findings against the visual trend of a widening gap. This dual-phase verification ensures robustness, culminating in the identification of 2019 as the peak disparity year. The model’s ability to integrate structured data processing, self-correction, and contextual alignment of quantitative and visual evidence underscores its proficiency in multimodal reasoning tasks.

\paragraph{Performance of the Preliminary Model}
We first evaluate the performance of the preliminary VLM (combined by ViT, MLP and a Qwen2.5-32B-Instruct) obtained after the initial MLP initialization step. This preliminary model achieves a competitive score of 64.0 on MMMU benchmark.

\paragraph{Performance of the Newly Assembled  Skywork R1V}
\label{sec:exp_model_reass}

Upon transferring and applying the pretrained MLP adapter to the DeepSeek-R1-distill-Qwen-32B model (i.e., after performed Model Re-Assembly), the newly assembled multimodal model achieves an impressive score of 60.2 (Table 
\ref{tab:iterative_sft_performance}). Remarkably, this performance not only exceeds many smaller-scale models explicitly trained for multimodal alignment but also rivals larger models such as InternVL2-40B (55.2). Moreover, the reassembled model's performance closely approaches state-of-the-art models at a similar scale, notably InternVL2.5-38B-MPO (64.1). These results illustrate a surprising effectiveness of the pretrained MLP in aligning the ViT vision encoder with another reasoning-capable LLM from the same series (DeepSeek-R1-distill-Qwen-32B), despite employing a different tokenizer and without additional fine-tuning.

\begin{table*}[t]
    
    \tablestyle{8pt}{1.1}
    \centering
    \small
        \begin{tabular}{@{}cccccccc@{}}
            \toprule
             \multirow{2}{*}{Model}& \multirow{2}{*}{Initial}& \multirow{2}{*}{Stage 1} & \multicolumn{4}{c}{Stage 2}&\multirow{2}{*}{Stage 3 (RL)}\\
             \cmidrule(rl){4-7}
             &  &  & $t$=1 &$t$=2&$t$=3 &$t$=4 &\\
            \midrule
            Performance & 60.2 & 62.5 & 63.9 & 64.7 & 65.2 & 65.6& 69.0 \\
            \bottomrule
        \end{tabular}
   \caption{Model Performance of Skywork R1V at Different Stages on the MMMU Dataset.}
    \label{tab:iterative_sft_performance}
\end{table*}

\paragraph{Effectiveness of Iterative SFT}
The iterative Supervised Fine-Tuning (SFT) strategy yields consistent performance improvements across successive training stages, as demonstrated in Table~\ref{tab:iterative_sft_performance}. Beginning from an initial score of 60.2, the model exhibits steady incremental gains at each iteration. By the conclusion of the fifth stage, performance reaches 65.6, clearly evidencing the effectiveness and stability of iterative fine-tuning in progressively refining model capabilities.

\paragraph{Effectiveness of RL Training}
The introduction of Group Relative Policy Optimization (GRPO) utilizing a ReLU-based reward function significantly boosts the model's performance, attaining an impressive increase to 69.0. This notable improvement underscores the efficacy of RL techniques in further enhancing multimodal reasoning capabilities.
Additionally, an intriguing observation emerged during the RL training phase: employing GRPO results in an increase in the length and detail of the model's outputs. This phenomenon aligns with previous observations from DeepSeekR1, where models undergoing RL training  exhibit "aha moments", spontaneously generating more comprehensive and elaborative responses.

\bibliography{main}
\bibliographystyle{plain}
\end{document}